\documentclass[conference]{IEEEtran}
\IEEEoverridecommandlockouts
\usepackage{cite}
\usepackage{amsmath,amssymb,amsfonts}
\usepackage{algorithmic}
\usepackage{graphicx}
\usepackage{textcomp}
\usepackage{xcolor}
\usepackage{numprint} 
\usepackage{caption}

\def\BibTeX{{\rm B\kern-.05em{\sc i\kern-.025em b}\kern-.08em
    T\kern-.1667em\lower.7ex\hbox{E}\kern-.125emX}}
\begin{document}

\title{Method for Comparison of Surrogate Safety Measures in Multi-Vehicle Scenarios
}

\author{\IEEEauthorblockN{Enrico Del Re}
\IEEEauthorblockA{\textit{Chair Sustainable Transport Logistics 4.0} \\
\textit{Johannes Kepler University Linz}\\
Linz, Austria \\
enrico.del\_re@jku.at}
\and
\IEEEauthorblockN{Cristina Olaverri-Monreal}
\IEEEauthorblockA{\textit{Chair Sustainable Transport Logistics 4.0} \\
\textit{Johannes Kepler University Linz}\\
Linz, Austria \\
cristina.olaverri-monreal@jku.at}
}

\maketitle

\begin{abstract}
With the race towards higher levels of automation in vehicles, it is imperative to guarantee the safety of all involved traffic participants. Yet, while high-risk traffic situations between two vehicles are well understood, traffic situations involving more vehicles lack the tools to be properly analyzed.
This paper proposes a method to compare Surrogate Safety Measures values in highway multi-vehicle traffic situations such as lane-changes that involve three vehicles. This method allows for a comprehensive statistical analysis and highlights how the safety distance between vehicles is shifted in favor of the traffic conflict between the leading vehicle and the lane-changing vehicle.

\end{abstract}


\section{Introduction}

In the quest to improve traffic safety, a significant challenge is the ability to differentiate between safe and unsafe situations. Initially, unsafe situations could only be identified by analyzing recorded accidents, but the frequency of these accidents in recorded datasets was too low to provide a comprehensive understanding. As a result, Traffic Control Techniques (TCTs) such as the Swedish TCT \cite{STCT_origin} have been developed to evaluate safety in non-accident scenarios.

These TCTs rely on a combination of the in-person observation by experts and measurement of parameters, such as Surrogate Safety Measures (SSMs)\cite{Laureshyn2018TheST}, to detect unsafe situations. These measures have been linked to accident rates and have been used to increase vehicle safety in conjunction with Advanced Driver Assist Systems (ADAS) \cite{1400969}. Vehicles equipped with such systems are considered Level 1 Autonomous Vehicles (AVs), and it is likely that they will also play an important role in higher-level AVs.

However, the relationship between SSMs, accident rates and human driving behavior is more complex. For example, the authors in \cite{DSSM} have shown that several SSMs do not fully match human driving acceleration and deceleration rates in car-following scenarios, and in \cite{OSA} also the limitations and challenges of SSMs with connected and automated vehicles and mixed traffic situations have been highlighted.
Additionally, human estimation and adherence to SSMs are not always accurate, further emphasizing the need for modifications.

Therefore, since autonomous vehicles should rely on and take into account human-like behavior \cite{smirnov2021game}, modifications or alternatives to currently used SSMs are necessary and research regarding their development is being performed \cite{MTTC_origin}. Most microscopic traffic safety research is focused on the interaction between only two vehicles, the ego vehicle and one vehicle it is in conflict with. As a result, SSMs are also typically defined regarding only one other traffic participant (or obstacle). In order to use SSMs in vehicles with high levels of automation, it is necessary to understand their role in traffic situations involving more vehicles and how human driving behavior differs from the optimal safe trajectory predictions that result from estimations using SSMs. 

Research on lane-change scenarios \cite{9987124} revealed that there is a statistically significant difference in SSM values in lane-changing maneuvers, depending on whether the ego vehicle is changing lanes in front of or behind another vehicle. Intuitively this result can be expected, though a quantitative analysis is necessary to take that behavior into account for AVs. In the past work however, these SSMs were mainly calculated between two vehicles only. Extending the observations to multi-vehicle scenarios lead to the research presented here.  

The focus of this paper is to propose a method to compare the SSM values of different traffic participants and apply it to lane-changes involving three vehicles on a highway.
We thus formulate the following hypothesis: When performing a lane-change between two vehicles, the safety distance measured by SSMs is identical between the lane-changing (ego vehicle) and leading vehicle, and the lane-changing and the following vehicle. After all, for an AV this could be the optimal trajectory. In section \ref{sec:Methodology} we propose a methodology that tests the defined hypothesis.

The next section presents a review of related literature, section \ref{sec:Data} the lane-change situation and dataset, and section \ref{sec:Methodology} a description of the applied methodology. The results of applying the method are shown in section \ref{sec:Results} and section \ref{sec:Conclusion} presents the conclusion and future research directions.

\section{Related Literature}

The groundwork for this paper comprises various research on SSMs, which can be found summarized in \cite{SSM}. Especially various modifications of SSMs such as extending them to consider more dynamic traffic situations \cite{MTTC_origin} or combining them to new ones \cite{Nadimi2021IntroducingNS}. However, to the authors' knowledge SSMs have remained focused on traffic situations involving only two vehicles.

One approach which does consider multi-vehicle conflicts has been presented in \cite{ARUN2023100252}. Instead of calculating SSM values between two vehicles, it defines a safety field for the ego vehicle where each position is influenced by the surrounding environment and vehicles. Still, attributing a safety value at each position requires the use of SSMs.

The factors affecting lane-change crashes specifically have also been analyzed in past research, though from a subjective approach \cite{factorslanechange}. Current regulations for the use of higher level AVs however are based on SSMs for safety assessment and decision making \cite{UN157}. An amendment to \cite{UN157} would allow an AV to perform a lane-change if it fulfills SSM-based safety thresholds with regards to the leading and following vehicle separately, splitting the lane-change into two separate traffic situations.

Thus, although there are existing works in the field of research, there is a research gap regarding the number of vehicles involved. Therefore we contribute in this paper to the body of knowledge and present a method to compare SSM values in highway lane-changes scenarios that involve three vehicles. This method allows for a comprehensive statistical analysis and highlights how the safety distance between vehicles is shifted in favor of the traffic conflict between the leading vehicle and the lane-changing vehicle.

\section{Studied road scenario and dataset description}
\label{sec:Data}
\subsection{Lane-change scenario}

\begin{figure}
    \centering
    \includegraphics[scale=0.4]{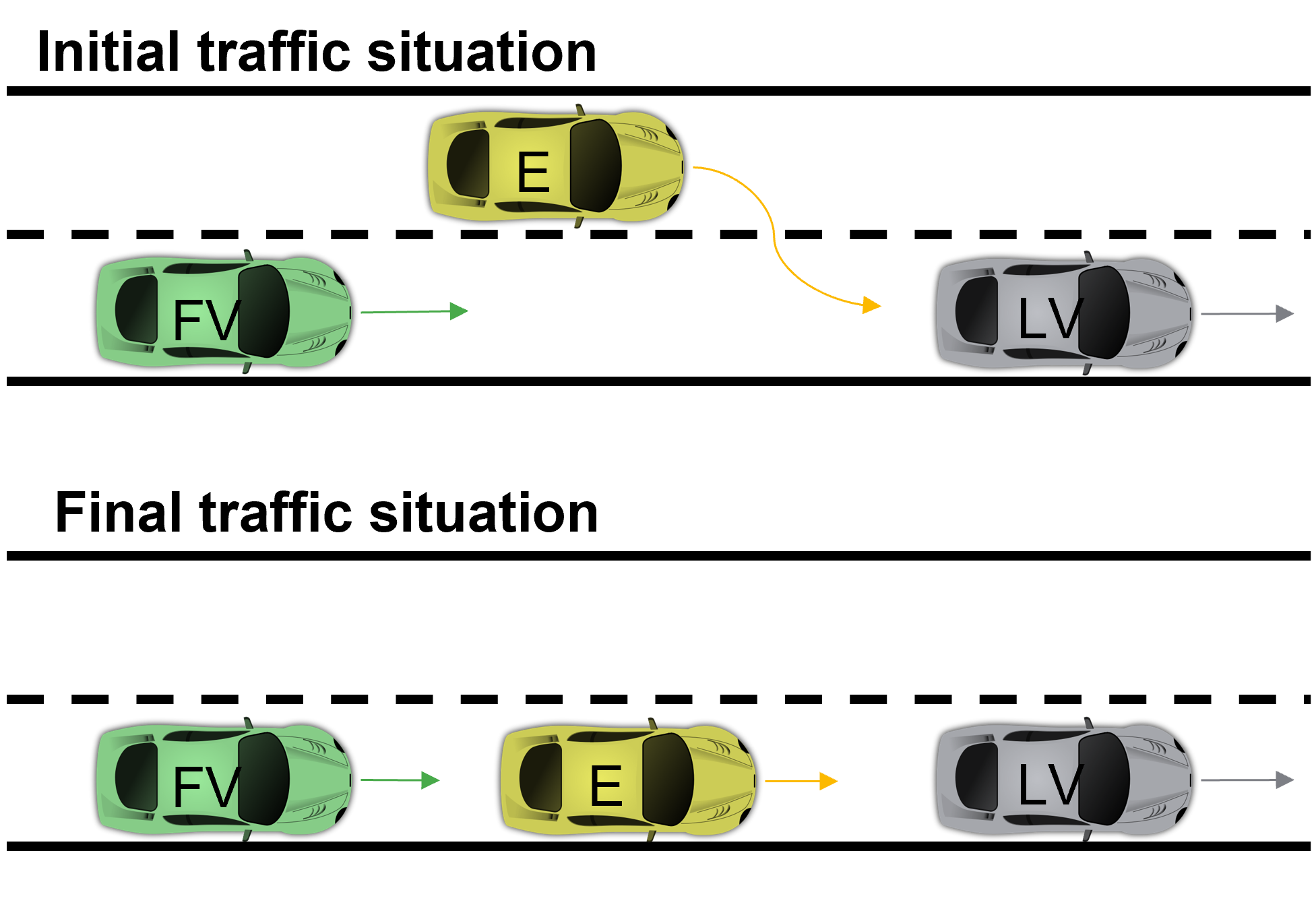}
    \caption{Lane-change to the adjacent lane by the vehicle E in between two vehicles, the leading vehicle (LV) and following vehicle (FV) of the final traffic situation. Visualized here for a lane-change to the right adjacent lane, though lane-changes to the left were also taken into consideration.} 
    \label{fig:Right_LC}
\end{figure}

The scenario chosen for comparing SSM values between multiple vehicles is a lane-change that is performed by an ego vehicle (E) in between a leading vehicle (LV) and a following vehicle(FV), as shown in Figure \ref{fig:Right_LC}. 

\subsection{Dataset}

The dataset used is part of the datasets recorded within the Next Generation SIMulation (NGSIM) program \cite{NGSIM}, specifically the dataset recorded on the Interstate 80 highway in California. The recorded area is approximately 500 meters long and consists of six freeway lanes, with one high-occupancy vehicle (HOV) lane and an additional lane on which vehicles drive onto the highway (on-ramp). In total 45 minutes of highway traffic between  4.00 p.m. and 5.30 p.m. were recorded.

This dataset was chosen since in (parts of) the US overtaking is allowed on both sides and could thus yield an equal number of lane-changes for both directions.


\section{Proposed Methodology}
\label{sec:Methodology}
\subsection{Surrogate Safety Measures}
\label{sec:M_SSMs}

This paper only uses a limited number of SSMs to assess traffic safety. However, the methodology for comparison introduced in Section \ref{sec:M_ratio} is suitable for a far larger number of SSMs and can be adapted if needed.

In general, SSMs can be grouped into proximity-based and deceleration-based categories, and they are most suitable for specific traffic situations. We limited ourselves to three distance-based (Time Headway, Inverse Time-to-Collision, Potential Index for Collision with Urgent Deceleration) and one deceleration-based SSM (minimum Deceleration Required to Avoid Crash). They are all well suited for car-following situations.
Their parameters, which are defined for interactions between two vehicles only are depicted in Table \ref{SSM_parameters}. In order to apply them to this work they will be calculated separately between the ego and the leading vehicle, as well as between the following and ego vehicle.

\begin{table}[htbp]
\caption{Parameters used to calculate SSMs}
\begin{center}
\begin{tabular}{| c | l |}
\hline
\textbf{Parameter} &\textbf{Definition} \\
\hline
 & distance between the front of the following \\ 
$D$ & vehicle and the rear part of the leading vehicle \\
\hline
$v_F$ & velocity of the following vehicle \\
\hline
$v_L$ & velocity of the leading vehicle  \\
\hline
$a$ & deceleration rate of both vehicles FV and LV for PICUD\\
\hline
 & reaction time of the following vehicle \\
$t_R$ & to a deceleration of the leading vehicle\\
\hline
\end{tabular}
\label{SSM_parameters}
\end{center}
\end{table}

\subsubsection{Time Headway (TH)}

Time Headway is defined as the time required for the following vehicle to reach the current position of the leading vehicle:
\begin{equation}
    TH = \frac{D}{v_F} \quad.
\end{equation}

A higher value indicates a safer traffic situation, with values above $2 [s]$ rarely seen in congested traffic \cite{TH_threshold}. This threshold is thus used to filter out free-flowing traffic as we are only interested in situations where interactions with other drivers have to be taken into account.

\subsubsection{Potential Index for Collision with Urgent Deceleration (PICUD)}

PICUD is defined as the distance between two vehicles if they come to a full stop after decelerating with the same force and the following vehicle is delayed by a reaction time. The deceleration rate and the reaction time are predefined parameters, in this case set to $a = 3.3 [m/s^2]$ and $t=1 [s]$ \cite{SSM}.

\begin{equation}
    PICUD = \frac{v_L^2-v_F^2}{2a}+D-v_Ft_R
\end{equation}

Again, a higher value indicates a safer traffic situation, though any value above 0 can be considered safe. Notably, negative values are possible in contrast with TH or the DRAC explained in the next paragraph.

\subsubsection{Minimum Deceleration Required to Avoid Crash (DRAC)}

DRAC indicates the deceleration rate required to avoid a collision, thus a lower value indicates a safer situation. If the two vehicles are not on a collision course the value is set to 0. 
\begin{equation}
    DRAC = \begin{cases}
    \frac{(v_F-v_L)^2}{D}, & v_F > v_L \\
    0, & else
    \end{cases}
\end{equation}

\subsubsection{Inverse Time-To-Collision (ITTC)}

While the standard Time-To-Collision (TTC) is more frequently used, it is undefined for non-collision trajectories. Though this is not an issue when calculating it between two vehicles, it becomes one when trying to compare TTC values with respect to both a leading and a following vehicle. Comparisons with two, or even one (as happens more frequently), undefined values are not possible.

Thus we are using in this work instead the inverse as defined in \cite{ITTC}:

\begin{equation}
    ITTC = \frac{v_F-v_L}{D}
\end{equation}

A lower value indicates a safer situation, with only values above 0 indicating a collision in $1/ITTC [s]$. Negative values instead increase with the acceleration and/or deceleration required to reach a collision course.

\subsection{Comparing SSMs}
\label{sec:M_ratio}

Since SSMs are used to evaluate the safety of any traffic situation, comparing two SSM values appears to be trivial, with their ratio being defined as:
\begin{equation}
\label{eqn:ratio_old}
    SSM_R = \frac{SSM_A}{SSM_B} \quad,
\end{equation}
where $SSM_A$ and $SSM_B$ are the measured values of a single SSM. 

\begin{figure}
    \centering
    \includegraphics[scale=0.8]{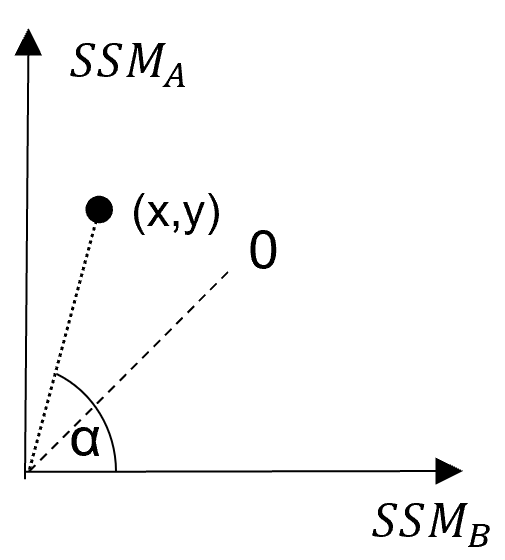}
    \caption{Visualization of the ratio transformation from equation \ref{eqn:ratio_old} with the angle $\alpha=\arctan(y/x)$ with $x$ the value of $SSM_B$ and $y$ the value of $SSM_A$. The dotted line marks the angle where the new ratio should take the value 0, at $\pi/4$.}
    \label{fig:Ratio_circle}
\end{figure}

For strictly positive- or negative-definite SSMs this is sufficient. However, PICUD, ITTC and DRAC amongst others do not fulfill this criterion. The information about the sign of $SSM_A$ and $SSM_B$ would be lost or it would be undefined for $SSM_B=0$.
Looking at the  ratio from equation \ref{eqn:ratio_old} in a 2-D plane (Figure\ref{fig:Ratio_circle}), we notice that it also  corresponds to $\tan \alpha$ for the point $(x,y)=(SSM_B,SSM_A)$.
As $\arctan (y/x) $ is defined for any real value of $y$ and $x$, we can thus extend equation \ref{eqn:ratio_old}.
For convenience we further choose to rescale the angle to $[-1,1]$, with the following conditions:

\begin{eqnarray}
    f(SSM_A, SSM_B): \mathbb{R}^+ \times \mathbb{R}^+ \rightarrow [-1,1]  \\
    f(x,x) = 0 \\
    f(0,x) = 1\\
    f(x,0) = -1,
\end{eqnarray}
which are fulfilled by 

\begin{equation}
    f_P(x,y)=-1+2\sin{\left (\arctan(y/x)\right )}.
\end{equation}

Extending the domain of definition to include negative numbers we get the following conditions: 
\begin{eqnarray}
\label{eq:ratio_R}
    f(SSM_A, SSM_B): \mathbb{R} \times \mathbb{R} \rightarrow [-1,1]  \\
    f(x,x) = 0 \\
    f(x,y) = -f(-x,-y) \\
    f(-x,x) = 1,\quad x>0 \\
    f(x,-x) = -1, \quad x>0, 
    \label{eq:ratio_R_2}
\end{eqnarray}
which are fulfilled by
\begin{equation}
    f_R(x,y) = \sin{\left ( \arctan(\frac{y}{x})- \frac{\pi}{4} \right )}.
\end{equation}
Again, a negative $f_R(x,y)$ indicates that $x$ had a higher value than $y$, with the highest relative value achieved at $f_R(x,y)=-1$, and vice-versa for $f_R(x,y)$ positive or $f_R(x,y)=1$.

For both $f_P$ and $f_R$ it is assumed that a higher value of an SSM indicates a safer situation. In case this is inverted (e.g. DRAC), using the negative of the function instead is sufficient to keep the previous definition of $-1$ and $1$.

\begin{figure}
    \centering
    \includegraphics[scale=0.8]{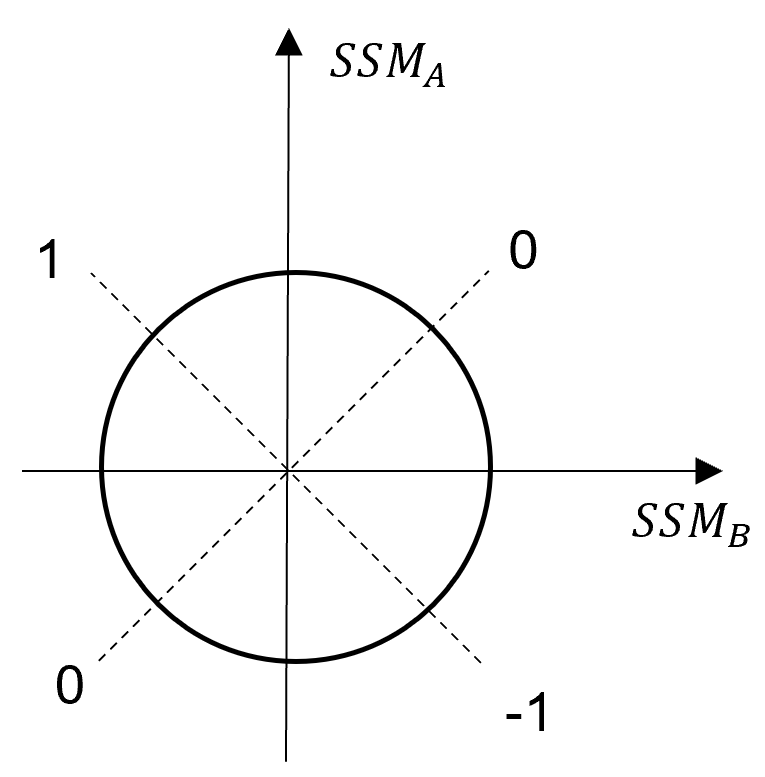}
    \caption{Visualization of the boundary conditions for an extension of the domain of the ratio between two SSMs to $\mathbb{R}$, defined in equations \ref{eq:ratio_R}-\ref{eq:ratio_R_2}. The dotted lines mark the angle where the ratio has to take the values $0$,$1$ and $-1$.}
    \label{fig:PICUD_transformation}
\end{figure}
Within this paper $SSM_A(y)$ is the SSM value calculated between the ego and the leading vehicle and $SSM_B(x)$ between the ego and the following vehicle. Thus a ratio of $-1$ indicates that the ego vehicle has kept a higher safety distance to the following vehicle than to the leading one. A $0$ indicates an even split in the safety distance between the leading and following vehicles.
The function $f_P(x,y)$ is used to compare TH and DRAC (image domain $\mathbb{R}^+$), whereas $f_R$ is used for ITTC and PICUD (image domain $\mathbb{R}$).

\subsection{Statistical Methods}
\label{Stats}

The method to compare SSMs introduced above allows us to perform statistical tests on the null hypothesis mentioned in the introduction, that the distribution of the ratio is symmetric around $0$. As the distributions of the ratio of SSMs differ from a normal distribution we have to rely on non-parametric tests.

A Wilcoxon signed-rank test is used to analyze the median of the distributions and thus the null hypothesis. The alternative hypothesis is that the ratio is centered around a higher value.
The statistical value of the test corresponds to the sum of ranks where the ratio is above $0$, the corresponding p-value indicates the likelihood to obtain a statistical value as large or larger with an underlying distribution centered around $0$.

To analyze differences between lanes and direction a Kruskal-Wallis H-test is used.
The null hypothesis for this test is a lack of differences between the various groups (especially the median). It calculates the test statistic $H$ with the sum of the average rank of each group. The corresponding p-value is obtained with $H$ having a hypothetical $\chi^2$ distribution. It is the survival function of the $\chi^2$ distribution evaluated at $H$. Should the null hypothesis be rejected, a Dunn’s Multiple Comparison Test is used as post hoc test to analyze which groups are different.

A Spearman R test is used to analyze the correlation between $SSM_R$s and traffic situations, in order to understand whether differences observed with the above tests are intrinsic properties of a lane-changes or consequences of external circumstances such as the velocities of the leading, following and ego vehicle.

The tests are all performed with a $5\%$ confidence level.

\section{Results}
\label{sec:Results}
\subsection{Data}
Extracting all lane-changes from the dataset, limiting the time headway between the leading vehicle and ego, as well as between the following vehicle and ego to less than $2 [s]$ resulted in 320 lane-changes. The only type of vehicles considered here were cars.

Further, eliminating lane-changes involving the HOV and on-ramp reduced the number of lane-changes to 199, which were further distributed as follows in Table \ref{Lane_change situations}.
Lane 2 only includes left lane-changes, i.e. lane-changes from lane 3 to lane 2, since lane 1 is the HOV lane. Similarly, lane 6 only includes right lane-changes as lane 7 is the on-ramp.

\begin{table}[htbp]
\caption{Lane-change scenarios from the NGSIM I-80 dataset used for the analysis}
\begin{center}
\begin{tabular}{| c | c | c | c |}
\hline
\textbf{Lane} &\multicolumn{3}{|c|}{\textbf{Direction}} \\
\cline{2-4} 
& Right & Left & \textbf{Total}\\
\hline
2 & 0 & 38 &40\\
\hline
3 & 4 & 27 &31\\
\hline
4 & 7 & 44 & 51\\
\hline
5 & 6 & 59 & 65\\
\hline
6 & 14 & 0 & 14\\
\hline
\textbf{Total} & \textbf{31} & \textbf{168} & \textbf{199} \\
\hline

\end{tabular}
\label{Lane_change situations}
\end{center}
\end{table}

\subsection{Ratio of SSMs}

\begin{figure}
    \centering
    \includegraphics[scale=0.5]{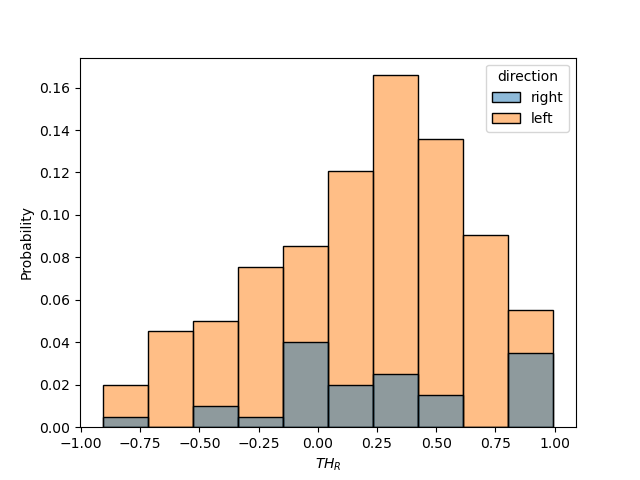}
    \caption{Histogram of the $TH_{R}$ visualized for different lane-change directions.}
    \label{fig:TH_ratio_all}
\end{figure}

\begin{figure}
    \centering
    \includegraphics[scale=0.5]{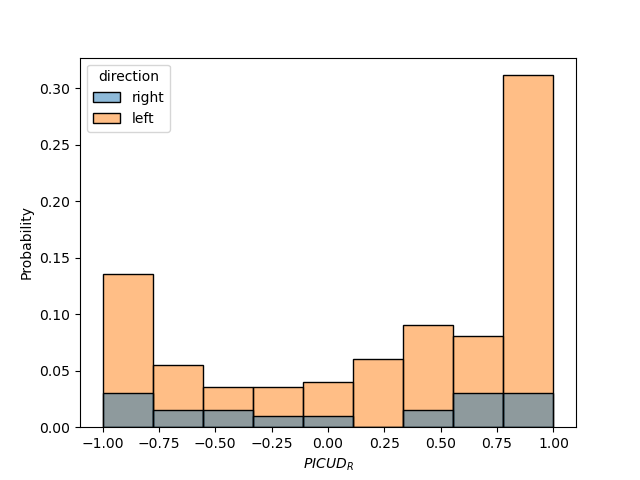}
    \caption{Histogram of the $PICUD_{R}$ visualized for different lane-change directions.}
    \label{fig:PICUD_ratio_all}
\end{figure}

\begin{figure}
    \centering
    \includegraphics[scale=0.5]{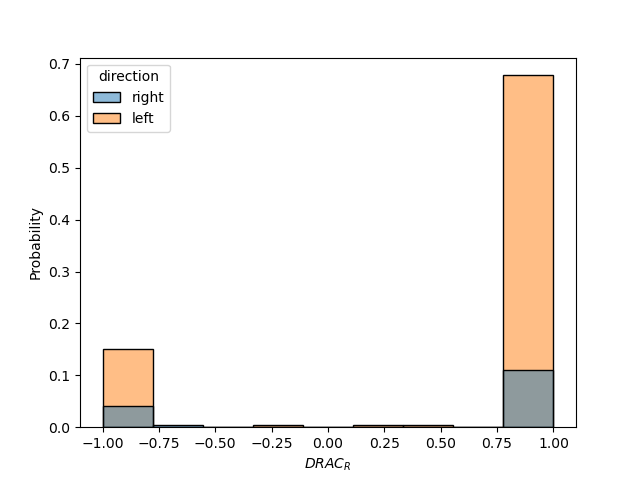}
    \caption{Histogram of the $DRAC_{R}$ visualized for different lane-change directions.}
    \label{fig:DRAC_ratio_all}
\end{figure}

\begin{figure}
    \centering
    \includegraphics[scale=0.5]{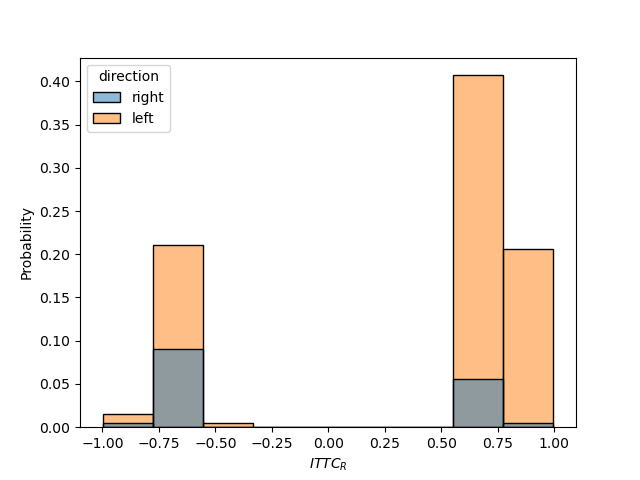}
    \caption{Histogram of the $ITTC_{R}$ visualized for different lane-change directions.}
    \label{fig:ITTC_ratio_all}
\end{figure}

For TH, PICUD and ITTC their ratios $TH_R$, $PICUD_R$ and $ITTC_R$ show a distribution between $[-1,1]$, as can be seen in Figure \ref{fig:TH_ratio_all}, Figure \ref{fig:PICUD_ratio_all} and Figure \ref{fig:ITTC_ratio_all}. For DRAC however, in most lane-change situations (188) the ego vehicle is on a collision course with only one of the vehicles, resulting in the $DRAC_R=-1$ or $DRAC_R=1$ values, as shown in Figure \ref{fig:DRAC_ratio_all}.

The results from the Wilcoxon signed-rank test with regards to the defined null and alternative hypothesis of a higher median are shown in Table \ref{W_rank_test}. The null hypothesis is rejected in all cases in favor of the alternative.

\npdecimalsign{.}
\nprounddigits{3}
\begin{table}[htbp]
\caption{Wilcoxon signed rank test for overall SSM ratios. W is the Wilcoxon test statistic with its corresponding p-value.}
\begin{center}
\begin{tabular}{| c | c | c |}
\hline
\textbf{SSM} & W & p-value \\
\hline
$TH_R$ & 14918 & 5.06e-10\\
\hline
$PICUD_R$ & 12945 & 1.15e-4\\
\hline
$DRAC_R$ &  16470 & 9.97e-20\\
\hline
$ITTC_R$ &  15948 & 8.29e-14\\
\hline
\end{tabular}
\label{W_rank_test}
\end{center}
\end{table}

All SSM ratios indicate that a higher safety distance has been kept with the leading vehicle than with the following one. 
In particular, $DRAC_R$ shows that collision courses with only the following vehicle are by far the more prevalent situation, 156 cases versus 32.

The results of the Spearman R test are shown in Table \ref{Spearman_R} and indicate either statistically insignificant or no correlations.

\npdecimalsign{.}
\nprounddigits{3}
\begin{table}[htbp]
\caption{Spearman R test to test the influence of velocities on SSM ratios. Correlation and statistical significance values are shown for the velocity of each vehicle.}
\begin{center}
\setlength\tabcolsep{0.5pt}
\begin{tabular}{| c | n{2}{5}| n{2}{5}|n{2}{5}| n{2}{5}|n{2}{5}| n{2}{5}| }
\hline
  & \multicolumn{2}{c|}{$v_{ego}$}& \multicolumn{2}{c|}{$v_{LV}$}& \multicolumn{2}{c|}{$v_{FV}$} \\
\cline{2-7}
\textbf{SSM} & \textbf{corr.} & \textbf{p-value}& \textbf{corr.} & \textbf{p-value}& \textbf{corr.} & \textbf{p-value} \\
\hline
$TH_R$ & 0.0585823723721705 & 0.4111303010389328 & 0.02776762600883204& 0.6970409058378425 &0.046852444038373686 & 0.5110957822877924 \\
\hline
$PICUD_R$ & 0.3347201934525483 & 0 & 0.12270290848180294& 0.08424607603382729 &0.046852444038373686 & 0.5110957822877924 \\
\hline
$DRAC_R$ & 0.414984783265845765 & 0 & 0.1173608045877719&0.09876751343166058 &0.046852444038373686 & 0.5110957822877924 \\
\hline
$ITTC_R$ & 0.5272162256926984 & 0 & 0.002014618547281864& 0.9774701870532531 &0.11361961321760317 & 0.11006891244143477 \\
\hline
\end{tabular}
\label{Spearman_R}
\end{center}
\end{table}

For lane and direction, the results of the Kruskal-Wallis H-test are shown in Table \ref{KW_lane} and show a difference for $TH_R$ and $DRAC_R$ when separated into lanes. 

\npdecimalsign{.}
\nprounddigits{3}
\begin{table}[htbp]
\caption{Kruskal-Wallis H-test for different lanes, with $H$ the test statistic and p-value the survival function of the $chi^2$ distribution at $H$.}
\begin{center}
\begin{tabular}{| c | n{2}{5}| n{2}{5}|}
\hline
\textbf{SSM} & \textbf{H} & \textbf{p-value}\\
\hline
$TH_R$ & 8.52552940745204 & 0.0741171591517968 \\
\hline
$PICUD_R$ & 14.004658327420316 & 0.007280203099052684 \\
\hline
$DRAC_R$ & 6.066186467981345& 0.19425965235597878 \\
\hline
$ITTC_R$ & 12.462285473205384 & 0.014225122917383563 \\
\hline
\end{tabular}
\label{KW_lane}
\end{center}
\end{table}

Using Dunn’s Multiple Comparison Test as post hoc test yields a significant difference between all lanes for $DRAC_R$ and for nearly all for $TH_R$, with the exception of the third and fifth lane.

For the direction of the lane-change the Kruskal-Wallis H-test resulted in a significant difference of the median only for $ITTC_R$, as depicted in Table \ref{KW_direction}.

\npdecimalsign{.}
\nprounddigits{3}
\begin{table}[htbp]
\caption{Kruskal-Wallis H-test for different lane-change direction, with $H$ the test statistic and p-value the survival function of the $chi^2$ distribution at $H$}
\begin{center}
\begin{tabular}{| c | n{2}{5}| n{2}{5}|}
\hline
\textbf{SSM} & \textbf{H} & \textbf{p-value}\\
\hline
$TH_R$ & 0.0755875576037397 & 0.7833685473867442 \\
\hline
$PICUD_R$ & 0.8029493087557285 & 0.37021303502492386 \\
\hline
$DRAC_R$ & 1.4685979418582908& 0.22556703722461036 \\
\hline
$ITTC_R$ & 13.562500000000114 & 0.00023074954964083063 \\
\hline
\end{tabular}
\label{KW_direction}
\end{center}
\end{table}

The combination of different lanes and lane-change direction was only possible for vehicles changing towards the left, due to a low number of right-changing cases. The results are shown in Table \ref{KW_left_lane} and indicate a difference in the median for $PICUD_R$ only. The  Dunn’s Multiple Comparison Test post hoc test showed a difference for all lanes.

\npdecimalsign{.}
\nprounddigits{3}
\begin{table}[htbp]
\caption{Kruskal-Wallis H-test for left changing vehicles split into different lanes, with $H$ the test statistic and p-value the survival function of the $chi^2$ distribution at $H$}
\begin{center}
\begin{tabular}{| c | n{2}{5}| n{2}{5}|}
\hline
\textbf{SSM} & \textbf{H} & \textbf{p-value}\\
\hline
$TH_R$ & 7.465286534755705 & 0.058457129975731964 \\
\hline
$PICUD_R$ & 13.69264990424847 & 0.003354811205488093 \\
\hline
$DRAC_R$ & 3.723931241834033 & 0.29285904999280366 \\
\hline
$ITTC_R$ & 6.602712213790937 & 0.08569862032878196 \\
\hline
\end{tabular}
\label{KW_left_lane}
\end{center}
\end{table}

Therefore, it is necessary to analyze the mean for individual lanes and directions as well. The Wilcoxon test results are shown in Table \ref{W_left_lane}. Only for the $5^{th}$ lane was the hypothesis of a median at $0$ accepted. Everywhere else this null hypothesis was rejected in favor of the alternative of being a median higher than $0$.

\npdecimalsign{.}
\nprounddigits{3}
\begin{table}[htbp]
\caption{Wilcoxon rank test for left changing vehicles split into different lanes for each $SSM_R$ with the test statistic W}
\begin{center}
\begin{tabular}{| c | c |n{2}{5}|}
\hline
\multicolumn{3}{|c|}{$TH_R$} \\
\hline
\textbf{Lane} & \textbf{W} & \textbf{p-value}\\
\hline
$2$ & 582 & 0.0003526316271973612 \\
\hline
$3$ & 330 & 0.0003526316271973612 \\
\hline
$4$ & 797 & 0.00021222297129865923 \\
\hline
$5$ & 1070 & 0.08130071611813722 \\
\hline

\multicolumn{3}{|c|}{$PICUD_R$} \\
\hline
\textbf{Lane} & \textbf{W} & \textbf{p-value}\\
\hline
$2$ & 586 & 0.0008882989962719475 \\
\hline
$3$ & 308 & 0.002125062183838671 \\
\hline
$4$ & 651 & 0.034337641488415234 \\
\hline
$5$ & 842 & 0.6272440704298268 \\

\hline
\multicolumn{3}{|c|}{$DRAC_R$} \\
\hline
\textbf{Lane} & \textbf{W} & \textbf{p-value}\\
\hline
$2$ & 658 & 1.4767707340532819e-06 \\
\hline
$3$ & 326.5 & 0.00014892154569324883 \\
\hline
$4$ & 893 & 8.658367801244551e-08 \\
\hline
$5$ & 1355.5 & 3.283570175174748e-05 \\

\hline
\multicolumn{3}{|c|}{$ITTC_R$} \\
\hline
\textbf{Lane} & \textbf{W} & \textbf{p-value}\\
\hline
$2$ & 677 & 4.395666770206776e-06 \\
\hline
$3$ & 326 & 0.0004984062045524902 \\
\hline
$4$ & 895 & 1.5201596083528361e-06 \\
\hline
$5$ & 1263 & 0.0021645118898306163 \\
\hline
\end{tabular}
\label{W_left_lane}
\end{center}
\end{table}

\section{Conclusion and Outlook}
\label{sec:Conclusion}

A new method to compare SSMs in multi-vehicle traffic situations was presented and applied to lane-change scenarios. It has allowed a more extensive statistical analysis which was previously impossible due to the variety of SSMs' image domains.

Vehicles changing lanes between two other vehicles have shown a strong preference towards keeping a higher safety distance towards the leading vehicle than the following one. The lane where the lane-change occurs has an impact on the median for DRAC, TH and ITTC, though it stays above $0$ for all SSMs and nearly all lanes.
However, the velocity of the vehicles itself has no significant impact on the ratio of the SSMs, thus not yielding a conclusive reason for the differences between lanes. Unfortunately, the limited number of lane-changes in the dataset also limited the statistical options, particularly when comparing the directions of the lane-change.

For this reason and to have more controllable surrounding conditions as well as more accurate data on the vehicles, a simulation with human participants is necessary. The resulting model for lane-changes will then be tested with participants to assess whether it accurately captured their safety concerns and would be acceptable for an AV. 

\section*{ACKNOWLEDGEMENT}

This work was supported by the Austrian Science Fund
(FWF), project number P 34485-N. It was additionally supported
by the Austrian Ministry for Climate Action, Environment,
Energy, Mobility, Innovation, and Technology (BMK)
Endowed Professorship for Sustainable Transport Logistics
4.0., IAV France S.A.S.U., IAV GmbH, Austrian Post AG and
the UAS Technikum Wien

\bibliographystyle{IEEEtran}
\bibliography{bliblio}

\begin{thebibliography}{10}
\providecommand{\url}[1]{#1}
\csname url@samestyle\endcsname
\providecommand{\newblock}{\relax}
\providecommand{\bibinfo}[2]{#2}
\providecommand{\BIBentrySTDinterwordspacing}{\spaceskip=0pt\relax}
\providecommand{\BIBentryALTinterwordstretchfactor}{4}
\providecommand{\BIBentryALTinterwordspacing}{\spaceskip=\fontdimen2\font plus
\BIBentryALTinterwordstretchfactor\fontdimen3\font minus
  \fontdimen4\font\relax}
\providecommand{\BIBforeignlanguage}[2]{{%
\expandafter\ifx\csname l@#1\endcsname\relax
\typeout{** WARNING: IEEEtran.bst: No hyphenation pattern has been}%
\typeout{** loaded for the language `#1'. Using the pattern for}%
\typeout{** the default language instead.}%
\else
\language=\csname l@#1\endcsname
\fi
#2}}
\providecommand{\BIBdecl}{\relax}
\BIBdecl

\bibitem{STCT_origin}
C.~Hydén and L.~Linderholm, ``The swedish traffic-conflicts technique,''
  \emph{International Calibration Study of Traffic Conflict Techniques}, pp.
  133--139, 01 1984.

\bibitem{Laureshyn2018TheST}
A.~Laureshyn and A.~V{\'a}rhelyi, ``The swedish traffic conflict technique:
  observer's manual,'' 2018.

\bibitem{1400969}
M.~Lu, K.~Wevers, R.~van~der Heijden, and T.~Heijer, ``Adas applications for
  improving traffic safety,'' in \emph{2004 IEEE International Conference on
  Systems, Man and Cybernetics (IEEE Cat. No.04CH37583)}, vol.~4, 2004, pp.
  3995--4002 vol.4.

\bibitem{DSSM}
S.~Tak, S.~Kim, D.~Lee, and H.~Yeo, ``A comparison analysis of surrogate safety
  measures with car-following perspectives for advanced driver assistance
  system,'' \emph{Journal of Advanced Transportation}, vol. 2018, 11 2018.

\bibitem{OSA}
M.~Elli, J.~Wishart, S.~Como, and S.~Dhakshinamoorthy, ``Evaluation of
  operational safety assessment (osa) metrics for automated vehicles in
  simulation,'' 04 2021.

\bibitem{smirnov2021game}
N.~Smirnov, Y.~Liu, A.~Validi, W.~Morales-Alvarez, and C.~Olaverri-Monreal, ``A
  game theory-based approach for modeling autonomous vehicle behavior in
  congested, urban lane-changing scenarios,'' \emph{Sensors}, vol.~21, no.~4,
  p. 1523, 2021.

\bibitem{MTTC_origin}
K.~Ozbay, H.~Yang, B.~Bartin, and S.~Mudigonda, ``Derivation and validation of
  new simulation-based surrogate safety measure,'' \emph{Transportation
  Research Record}, vol. 2083, no.~1, pp. 105--113, 2008.

\bibitem{9987124}
E.~del Re and C.~Olaverri-Monreal, ``Implementation of road safety perception
  in autonomous vehicles in a lane change scenario,'' in \emph{2022 IEEE
  International Conference on Vehicular Electronics and Safety (ICVES)}, 2022,
  pp. 1--6.

\bibitem{SSM}
S.~M. Mahmud, L.~Ferreira, M.~Hoque, and A.~Hojati, ``Application of proximal
  surrogate indicators for safety evaluation: A review of recent developments
  and research needs,'' \emph{IATSS Research}, vol.~41, 03 2017.

\bibitem{Nadimi2021IntroducingNS}
N.~Nadimi, A.~M. Amiri, and A.~Sadri, ``Introducing novel statistical-based
  method of screening and combining currently well-known surrogate safety
  measures,'' \emph{Transportation Letters}, vol.~0, no.~0, pp. 1--11, 2021.

\bibitem{ARUN2023100252}
\BIBentryALTinterwordspacing
A.~Arun, M.~M. Haque, S.~Washington, and F.~Mannering, ``A physics-informed
  road user safety field theory for traffic safety assessments applying
  artificial intelligence-based video analytics,'' \emph{Analytic Methods in
  Accident Research}, vol.~37, p. 100252, 2023. [Online]. Available:
  \url{https://www.sciencedirect.com/science/article/pii/S2213665722000410}
\BIBentrySTDinterwordspacing

\bibitem{factorslanechange}
\BIBentryALTinterwordspacing
M.~Shawky, ``Factors affecting lane change crashes,'' \emph{IATSS Research},
  vol.~44, no.~2, pp. 155--161, 2020. [Online]. Available:
  \url{https://www.sciencedirect.com/science/article/pii/S0386111219300020}
\BIBentrySTDinterwordspacing

\bibitem{UN157}
``Un regulation no 157 – uniform provisions concerning the approval of
  vehicles with regards to automated lane keeping systems [2021/389],'' pp.
  75--137, Mar 2021.

\bibitem{NGSIM}
``U.s. department of transportation federal highway administration. (2016).
  next generation simulation (ngsim) vehicle trajectories and supporting
  data,'' accessed: 2022-12-13.

\bibitem{TH_threshold}
T.~Ayres, L.~Li, D.~Schleuning, and Y.~Douglas, ``Preferred time-headway of
  highway drivers,'' 02 2001, pp. 826 -- 829.

\bibitem{ITTC}
V.~E. Balas and M.~M. Balas, ``Driver assisting by inverse time to collision,''
  \emph{2006 World Automation Congress}, pp. 1--6, 2006.

\end{thebibliography}

\end{document}